\documentclass[10pt,twocolumn,letterpaper]{article}

\usepackage[pagenumbers]{iccv} 

\definecolor{iccvblue}{rgb}{0.21,0.49,0.74}
\PassOptionsToPackage{hyphens}{url}
\usepackage[pagebackref,breaklinks,colorlinks,allcolors=iccvblue]{hyperref}

\makeatletter\renewcommand\paragraph{\@startsection{paragraph}{4}{\z@}
  {.5em \@plus1ex \@minus.2ex}{-.5em}{\normalfont\normalsize\bfseries}}\makeatother
\def\confName{ICCV}
\def\confYear{2025}

\newcommand{\name}{GroundingSuite}
\newcommand{\gseval}{{GSEval}}
\newcommand{\gstrain}{{GSTrain-10M}}

\title{GroundingSuite: Measuring Complex Multi-Granular Pixel Grounding}

\author{
    Rui Hu$^{ 1*}$ \quad
    Lianghui Zhu$^{ 1*}$ \quad
    Yuxuan Zhang$^{ 1}$ \quad
    Tianheng Cheng$^{ 1\dagger}$\quad
    Lei Liu$^{ 2}$ \quad
    Heng Liu$^{ 2}$\\
    Longjin Ran$^{ 2}$ \quad
    Xiaoxin Chen$^{ 2}$ \quad
    Wenyu Liu$^{ 1}$ \quad
    Xinggang Wang$^{ 1\ddagger}$\\
    $^{*}$ Equal contribution\quad$^{\dagger}$ Project lead\quad$^{\ddagger}$ Corresponding author\\
    $^1$ School of EIC, Huazhong University of Science \& Technology \\
    $^2$ vivo AI Lab\\
    Codes are available at: \href{https://github.com/hustvl/GroundingSuite}{\textcolor{magenta}{\texttt{hustvl/GroundingSuite}}}
}

\begin{document}

\makeatletter

\g@addto@macro\@maketitle{
\begin{figure}[H]
    \setlength{\linewidth}{\textwidth}
    \setlength{\hsize}{\textwidth}
    \vspace{-10mm}
    \centering
    \includegraphics[width=1\linewidth]{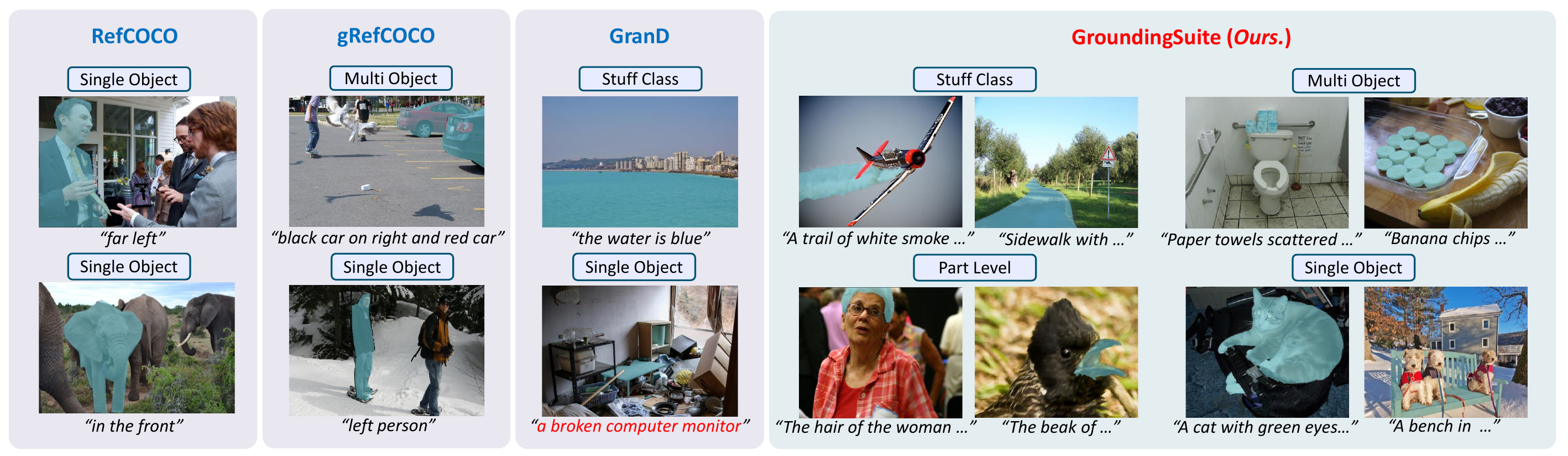}
  \vspace{-18pt}
  \caption{\textbf{Examples of our GroundingSuite dataset.} Including four aspects: stuff-class segmentation for context-aware localization, part-level segmentation requiring fine-grained understanding, multi-object segmentation with complex referential relationships, and single-object segmentation across diverse appearance variations. Note that `\textcolor{red}{red}' color means the referring is wrong.}

    \label{fig:example}
\end{figure}
}

\maketitle

\begin{abstract}
Pixel grounding, encompassing tasks such as Referring Expression Segmentation (RES), has garnered considerable attention due to its immense potential for bridging the gap between vision and language modalities.
However, advancements in this domain are currently constrained by limitations inherent in existing datasets, including limited object categories, insufficient textual diversity, and a scarcity of high-quality annotations. 
To mitigate these limitations, we introduce \textbf{GroundingSuite}, which comprises: 
(1) an automated data annotation framework leveraging multiple Vision-Language Model (VLM) agents; 
(2) a large-scale training dataset encompassing 9.56 million diverse referring expressions and their corresponding segmentations; 
and (3) a meticulously curated evaluation benchmark consisting of 3,800 images. 
The GroundingSuite training dataset facilitates substantial performance improvements, enabling models trained on it to achieve state-of-the-art results. Specifically, a cIoU of 68.9 on gRefCOCO and a gIoU of 55.3 on RefCOCOm. 
Moreover, the GroundingSuite annotation framework demonstrates superior efficiency compared to the current leading data annotation method, i.e., $4.5 \times$ faster than the GLaMM. 
\end{abstract}    
\section{Introduction}
\label{sec:intro}

Recent advancements in pixel grounding have fostered significant interest due to their remarkable segmentation performance in following natural language descriptions.
However, this task remains constrained by existing benchmark limitations. 
As shown in Tab.~\ref{tab:1}, widely-used datasets like RefCOCO series~\cite{referit,refcoco, refcocog,refcocog2}, while valuable for early research, restrict their scope to 
closed-set categories and limited manually annotated data samples.
Specifically, 
only 80 object categories from the COCO dataset~\cite{coco} hinder the pixel grounding task from generalizing to open-vocabulary understanding, diverse granularity levels (\eg, part-level segmentation), and complex scene compositions (\eg, multi-object interactions and background separation). 
Besides, current datasets can not satisfy both the amount and annotation quality requirements, \ie, manually annotated datasets can not scale up due to the heavy human effort, and automatically annotated datasets often show inferior label quality.

Some automatic annotation approaches, such as GLaMM~\cite{GLaMM} and MRES~\cite{refcocom}, are proposed to alleviate the heavy human annotation burden but also face low-quality annotation and high-cost
utilization problems. 
From the perspective of annotation quality, GLaMM suffers from unresolved textual ambiguities, and MRES only works with restricted fixed vocabularies. 
As for the cost, GLaMM has 23 pipeline steps, requiring a large amount of GPU resources, and MRES relies on human-annotated boxes as initialization.

To address the above challenges,
we propose GSSculpt, an annotation framework that involves vision-language models (VLM) as efficient annotation agents and effective quality checkers.
Specifically, GSSculpt incorporates three critical components: entity spatial localization, grounding text generation, and noise filtering. 
Entity spatial localization is the first phase, in which we generate comprehensive image captions, employ precise phrase grounding techniques, and utilize SAM2~\cite{sam2} to extract high-quality masks.
Then, we use state-of-the-art multimodal models with carefully designed prompts to generate unambiguous descriptions with clear positional relationships in the grounding text generation phase.
Last, we employ instruction-based segmentation models to filter the noisy references, ensuring high data quality throughout the collection.

Based on the GSSculpt, we further curated a large-scale training set, \ie, GSTrain-10M, and a comprehensive evaluation benchmark, \ie, \gseval{}. 
The above three components make up the proposed \textbf{GroundingSuite}. 
Specifically, GSTrain-10M comprises a 9.56-million-image training corpus automatically annotated on SA-1B dataset images~\cite{sam} using our hybrid framework, 
featuring diverse textual descriptions averaging 16 words in length and unambiguous instruction-mask pairs.
\gseval{} is a novel evaluation benchmark containing 3,800 images carefully selected from COCO's unlabeled datasets~\cite{coco}, ensuring zero overlap with existing annotated sets while maintaining natural scene diversity. 
As shown in Fig.~\ref{fig:example}, the proposed benchmark specifically covers four mainstream aspects of segmentation: stuff-class segmentation~\cite{cocostuff,ade20k,ade2} for context-aware localization, part-level segmentation~\cite{pascal_part,liang2015human,he2022partimagenet} of fine-grained understanding, multi-object segmentation~\cite{grefcoco} with complex referential relationships, and single-object segmentation~\cite{referit, refcoco,refcocog,refcocog2} across diverse appearance variations.

\begin{table}[t]
\centering
\small
\setlength{\tabcolsep}{3.5pt}
\begin{tabular}{l|ccccccccc}
\toprule
\textbf{Benchmarks} & \textbf{Cat.} & \textbf{Len.} & \textbf{Stuff} & \textbf{Part} & \textbf{Multi} & \textbf{Single}\\ 
\midrule
RefCOCO~\cite{referit,refcoco} & 80 & 3.6 &  &  & & \checkmark\\
RefCOCO+~\cite{referit,refcoco} & 80 & 3.5 &  &  & & \checkmark\\
RefCOCOg~\cite{refcocog,refcocog2} & 80 & 8.4 &  &  & & \checkmark\\
gRefCOCO~\cite{grefcoco} & 80 & 3.7 &  &  & \checkmark & \checkmark \\
RefCOCOm~\cite{refcocom} & 471 & 5.1 &  & \checkmark  &  & \checkmark \\
\textbf{\gseval{}} & \textbf{$\infty$} & \textbf{16.1} & \checkmark & \checkmark & \checkmark & \checkmark\\
\bottomrule
\end{tabular}
\caption{\textbf{Comparisons with previous Referring Expression Segmentation benchmark.} `Cat.' and `Len.' denote the number of categories and average text length; Stuff: includes stuff classes; Part: includes part-level annotations; Multi: supports multi-object references; Single: supports single-object references.}
\vspace{-.1 in}
\label{tab:1}
\end{table}

Our main contributions of \name{} can be summarized as follows:
\begin{itemize}
    \item To address the low-quality annotation and high-cost utilization problems of existing auto-labeling methods, we propose GSSculpt, a vision-language models (VLM) based automatic annotation framework, which produces accurate annotations and reduces 78\% of pipeline steps when compared to GLaMM.
    \item We introduce a large-scale training dataset and a carefully curated evaluation benchmark, laying a solid foundation for future research in pixel grounding. This dataset addresses critical limitations in existing grounding datasets, \ie, limited object categories, insufficient textual diversity, and a scarcity of high-quality annotations, while supporting diverse segmentation scenarios.
    \item The proposed dataset presents substantial performance enhancements across baseline methods. Models trained on our data consistently demonstrate superior results, establishing new state-of-the-art benchmarks across multiple evaluation metrics. Specifically, our approach achieves a cIoU of 68.9 on gRefCOCO and a gIoU of 55.3 on RefCOCOm.
\end{itemize}
\section{Related Work}

\subsection{Automatic Annotation for Pixel Grounding}

 CLIP~\cite{clip} and GLIP~\cite{glip} pioneered using vision-language models for label generation, while SAM~\cite{sam} enabled promptable segmentation at scale. Building on these, recent work explores automatic dataset creation through model collaboration. GranD~\cite{GLaMM} is Developed using an automatic annotation pipeline and verification criteria, it encompasses 7.5M unique concepts grounded in 810M regions. However, the GLaMM~\cite{GLaMM} process has many steps, leading to the accumulation of errors from multiple models, and as a result, it hasn't solved the problem of text ambiguity. MRES~\cite{refcocom} build a large visual grounding dataset namely MRES32M, which comprises over 32.2M high-quality masks and captions on the provided 1M images. However, it relies on box annotations from existing datasets and only uses a fixed vocabulary, which limits its generality.

\subsection{Pixel Grounding Benchmarks}

 Mainstream benchmarks such as RefCOCO~\cite{referit,refcoco}, RefCOCO+~\cite{referit,refcoco}, and RefCOCOg~\cite{refcocog,refcocog2} have driven progress in language-guided segmentation. However, their reliance on COCO's categorical constraints (80 classes) and human-annotated referring expressions creates artificial domain limitations. 
 Among recent works, RefCOCOm~\cite{refcocom} contains 80 object categories and an associated 391 part categories. Despite its part-level advancements, RefCOCOm’s reliance on COCO’s fixed 80 object categories limits its ability to benchmark open-vocabulary or cross-category referring segmentation, hindering evaluation of models' adaptability to unseen object types in real-world scenarios. 
 GRES~\cite{grefcoco} is a new benchmark called Generalized Referring Expression Segmentation, which extends the classic RES to allow expressions to refer to an arbitrary number of target objects. The GRES benchmark has several limitations, including its restriction to the 80 object categories from the COCO dataset, the lack of part-level segmentation, and the absence of a background class for segmentation.

\section{GSSculpt: Large-scale Grounding Labeling}
\label{ref:gssculpt}
\begin{figure*}
\centering
\includegraphics[width=1.0\linewidth]{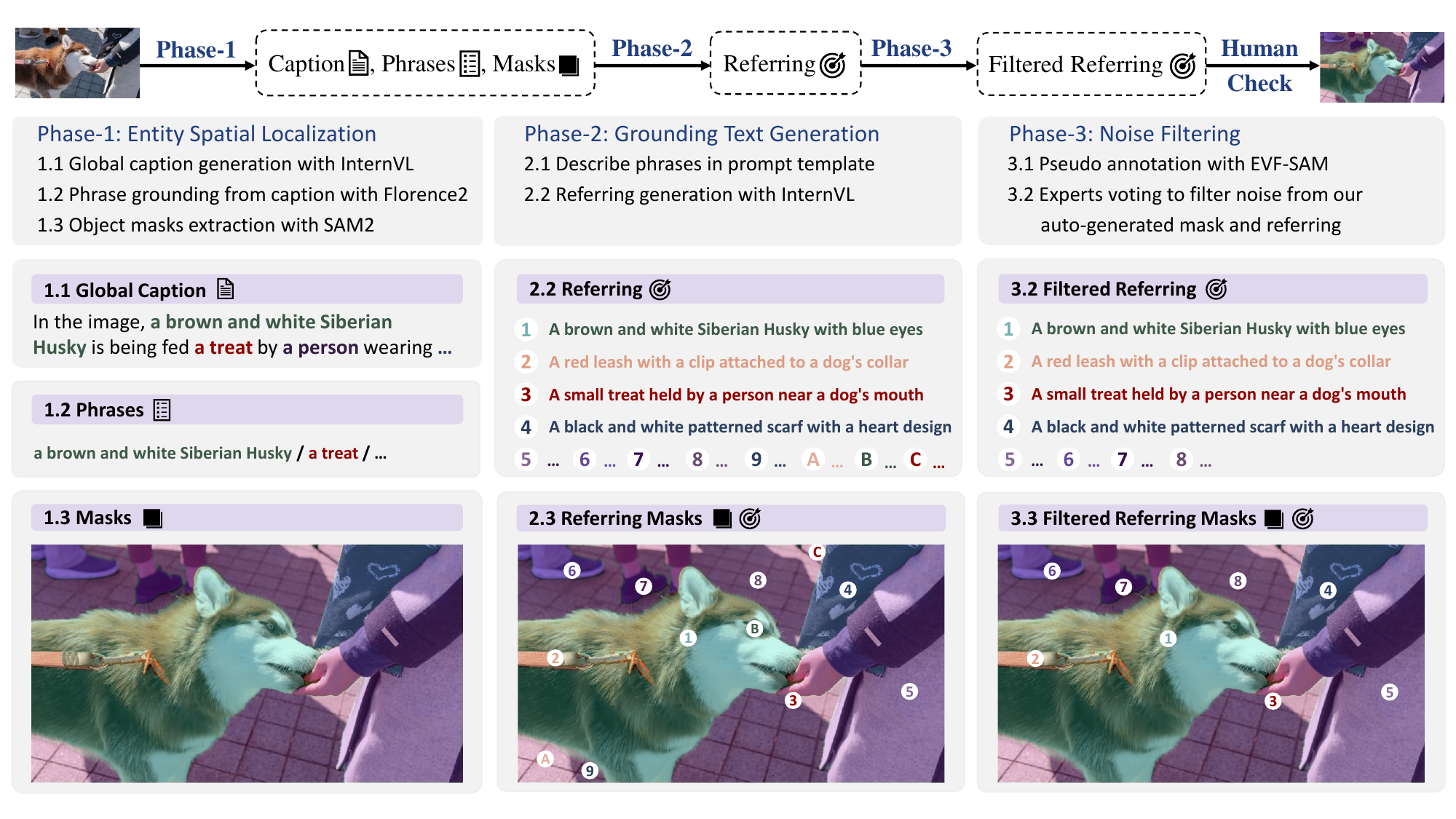}
\vspace{-20pt}
\caption{\textbf{GSScuplt Automatic Annotation Framework.} Our pipeline consists of three sequential phases: (1) Entity Spatial Localization, where we first identify potential objects of interest and generate high-quality segmentation masks; (2) Grounding Text Generation, where we then create unambiguous natural language descriptions that uniquely reference the segmented objects; and (3) Noise Filtering, where we finally eliminate ambiguous or low-quality samples to ensure dataset reliability.}
\vspace{-.2 in}
\label{fig:pipeline}
\end{figure*}

\subsection{Overview}

We introduce our vision-language models (VLM) based automatic annotation framework \textbf{GSSculpt}, 
designed to automatically generate high-quality pixel grounding data at scale.
Through a comprehensive analysis of existing approaches, we propose three critical components essential for the effective auto-labeling framework, as shown in Fig.~\ref{fig:pipeline}, \ie, (1) \textbf{entity spatial localization}: discovering regions/objects of interest and generating high-quality masks; (2) \textbf{grounding text generation}: generating precise and uniquely identifiable language descriptions for regions or objects; and (3) \textbf{noise filtering}: eliminating ambiguous or low-quality samples.
The proposed framework GSSculpt provides elaborate designs for each component to ensure annotation quality and accuracy, collectively constructing an efficient streamlined annotation framework, which aims to advance the state of pixel-level grounding datasets.

\paragraph{Data source.} 
For large-scale training data, we primarily utilize SA-1B~\cite{sam} as our data source, which comprises 1.1 billion high-quality segmentation masks across 11 million diverse, high-resolution images.
In this paper, we sample 2M images for annotation. While the segmentation annotations in SA-1B focus on diverse geometric prompts, we concentrate on semantically meaningful objects or regions. Therefore, we employed SAM-2~\cite{sam2} for segmentation annotation instead of directly using the mask annotations provided by SA-1B.

\subsection{Entity Spatial Localization}
The proposed framework builds upon precise object/region recognition and localization within complex visual scenes. This critical first stage employs a sophisticated three-tiered approach that systematically addresses the challenges of visual parsing.
\paragraph{Global caption generation.} We leverage a cutting-edge large visual-language model, \ie, InternVL2.5~\cite{internvl2.5}, to produce comprehensive scene descriptions, discovering all semantically significant objects or regions within each image with remarkable completeness.

\paragraph{Phrase grounding.} The generated captions serve as input for Florence-2~\cite{florence2}, as the advanced phrase grounding model, which precisely grounds texts to corresponding spatial locations. This process provides preliminary bounding box regions for candidate objects.

\paragraph{Mask generation.} Then we adopt SAM2~\cite{sam2} to obtain pixel-level segmentation masks for the grounded texts using bounding boxes as spatial prompts.

\subsection{Grounding Text Generation}
The second stage primarily further optimizes the textual descriptions of grounded regions with Large Language Models, enriching the grounding texts with the context of the image.

We design specialized prompt templates to guide multimodal language models in generating distinct and unambiguous references. These templates strategically emphasize spatial relationships, distinctive visual features, and contextual cues. Using powerful models like InternVL2.5, we generate linguistically diverse and rich descriptions. Our method produces natural descriptions averaging 16 words, significantly more expressive than the shorter phrases in manually annotated datasets, while maintaining referential clarity and eliminating ambiguity.

\subsection{Noise Filtering}
The noise filtering stage ensures dataset quality by eliminating ambiguous or incorrect annotations:

We employ instruction-based segmentation models, \ie, EVF-SAM~\cite{evfsam}, to identify potentially ambiguous referring expressions by measuring consistency between the generated expression and the corresponding mask. Specifically, we use the generated texts in previous steps to prompt a Referring Expression Segmentation (RES) model to produce masks, and then calculate the IoU between the generated masks and the annotated mask. 
By applying an IoU threshold, \ie, 0.5, we can effectively filter out inaccurate text-mask pairs.

This comprehensive filtering approach achieves an optimal balance between dataset scale and annotation quality, resulting in 9.56 million high-quality training samples while ensuring the integrity and quality of the final dataset.

\subsection{\gstrain{}}
Applying the proposed annotation framework to a diverse subset of images from the SA-1B dataset, we have created \textbf{\gstrain{}}, a large-scale comprehensive training dataset, aiming for pixel grounding and comprising 9.56 million high-quality text-mask pairs across 2 million images. \gstrain{} contains linguistically rich descriptions with an average length of 16 words, significantly more detailed than existing manually annotated datasets. The \gstrain{} dataset covers an open vocabulary of concepts including common objects, fine-grained parts, amorphous stuff categories, and diverse scenes. Each annotation undergoes a rigorous noise-filtering process, ensuring unambiguous references with clear spatial relationships. The presented \gstrain{} represents a substantial advancement in both scale and quality for pixel grounding, enabling more robust and generalizable model training across a wider range of visual concepts and linguistic expressions.
\vspace{-.09 in}

\section{\gseval: A Comprehensive Benchmark}
\begin{figure*}
\centering
\includegraphics[width=1.0\linewidth]{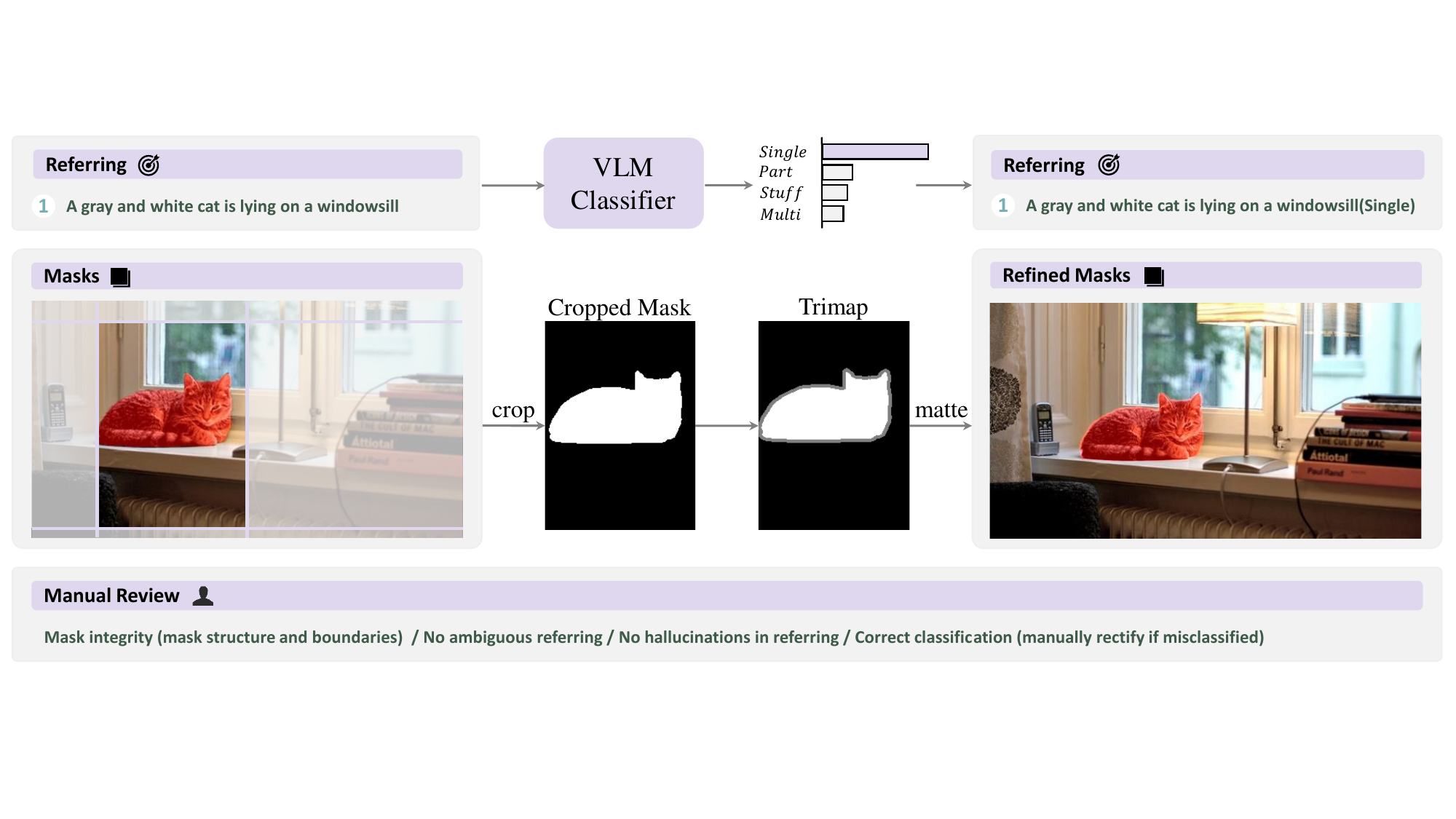}
\vspace{-15pt}
\caption{\textbf{Curation pipeline for GSEval Benchmark.} 
First, we apply our annotation pipeline to unlabeled COCO images. Next, we use a VLM classifier to ensure the categories of referring prompts. Then, we translate the coarse masks to trimaps and apply matting methods for precise object boundaries. Finally, we organize human reviewers for manual checks.
}
\vspace{-.2 in}
\label{fig:test-pipeline}
\end{figure*}

Building upon the aforementioned annotation framework, we further develop \textbf{\gseval{}}, a comprehensive evaluation benchmark for pixel grounding tasks. As shown in Fig.~\ref{fig:test-pipeline}, we propose a human-guided curation pipeline to ensure data quality and task diversity.

\subsection{Automatic Data Curation}

We first apply the proposed annotation framework (Sec.~\ref{ref:gssculpt}) to the unlabeled images of the COCO dataset~\cite{coco}.
The generated annotations and labeled images have no overlap with existing labeled datasets.
Then, we use vision-language models (VLM) to assign categories for referring prompts. This pre-categorization helps to reduce the cost of subsequent manual selection and filters out noisy referring-mask pairs.
Last, we adopt matting methods to refine the boundaries of object masks. Specifically, we translate the mask areas to trimaps and use the \textit{off-the-shelf} matting model~\cite{vitmatte} to generate precise boundaries.

\subsection{Human Data Curation}

To ensure the quality of ~\gseval{}, every image was manually selected and verified by human reviewers. This process resulted in a dataset organized into four distinct categories:
\begin{itemize}
    \item \textit{Stuff Segmentation}: Comprises 1,000 images of background elements that require context-aware localization, such as \texttt{sky} and \texttt{sea}.
    \item \textit{Part-level Segmentation}: Consists of 500 images that demand fine-grained understanding of object components, such as \texttt{the camera on a phone} or \texttt{a man's beard}.
    \item \textit{Multi-object Segmentation}: Includes 800 images with complex referential relationships between entities, like \texttt{a flock of sheep} or \texttt{two dogs}.
    \item \textit{Single-object Segmentation}: Contains 1,500 images that showcase diverse object appearances, such as \texttt{a brown cat} and \texttt{a colorful parrot}.
\end{itemize}

This enhanced human-in-the-loop approach significantly improves efficiency while maintaining strict quality standards.
The VLM pre-classification reduces the human reviewers' annotation burden, allowing them to focus on nuanced verification. 
By strategically distributing images across the four categories, we ensure comprehensive evaluation across varying levels of granularity and complexity. 
The curated benchmark covers multiple domains and object types, providing a robust assessment of models' generalized segmentation capabilities in real-world scenarios.

\subsection{\gseval-BBox}
In addition to the pixel-based segmentation benchmark, we construct a bounding box version, namely \textbf{\gseval{}-BBox}.
Specifically, \gseval{}-BBox is designed to evaluate the visual grounding capabilities of multimodal large language models. 
\gseval{}-BBox converts the high-quality segmentation masks into corresponding bounding boxes, enabling direct assessment of referential object localization.

\subsection{Evaluation Metrics}

\paragraph{Pixel-level grounding.}
We adopt the standard metric for pixel level evaluation, \ie, $\text{gIoU}$. The $\text{gIoU}$ metric calculates the average of per-image Intersection-over-Union (IoU) scores:
\vspace{-.1 in}
$$\text{gIoU} = \frac{1}{N}\sum_{i=1}^{N}\frac{|P_i \cap G_i|}{|P_i \cup G_i|},$$
where $P_i$ and $G_i$ denote the predicted segmentation mask and ground truth mask for the $i$-th sample, respectively. $N$ is the total number of samples.
In contrast to previous methods using cIoU, the proposed \gseval{} includes part-level segmentation tasks and we prioritize $\text{gIoU}$ as the primary evaluation metric. 
Specifically, $\text{gIoU}$ provides a more balanced assessment across objects of varying sizes, 
while $\text{cIoU}$ exhibits bias toward larger objects and demonstrates greater volatility in measurements.

\paragraph{Box-level grounding.}
For box-level grounding, we adopt the same metrics as in referring expression comprehension (REC), which typically includes accuracy at different IoU thresholds (\eg, Acc@0.5).

\section{Evaluation on \gseval}
In this section, we evaluate several representative methods on our proposed \gseval{} under zero-shot settings.

\subsection{Benchmark Details}
As shown in Tab.~\ref{tab:1}, \gseval{} significantly surpasses existing referring expression segmentation benchmarks in multiple dimensions. 
Compared to the RefCOCO dataset, which primarily relies on COCO category annotations, our proposed GSEval adopts an open-vocabulary setting, encompassing not only foreground objects but also stuff categories and various part-level categories.
In addition, the average text length in \gseval{} reaches 16.1 words, making its descriptions substantially more detailed and nuanced compared to others. Furthermore, \gseval{} is uniquely comprehensive in supporting all key features: stuff classes, part-level annotations, and both multi-object and single-object references. This comprehensive design enables more challenging and realistic evaluation scenarios that better reflect real-world language grounding applications.

\begin{figure*}[h]
\begin{center}
\includegraphics[width=1.0\linewidth]{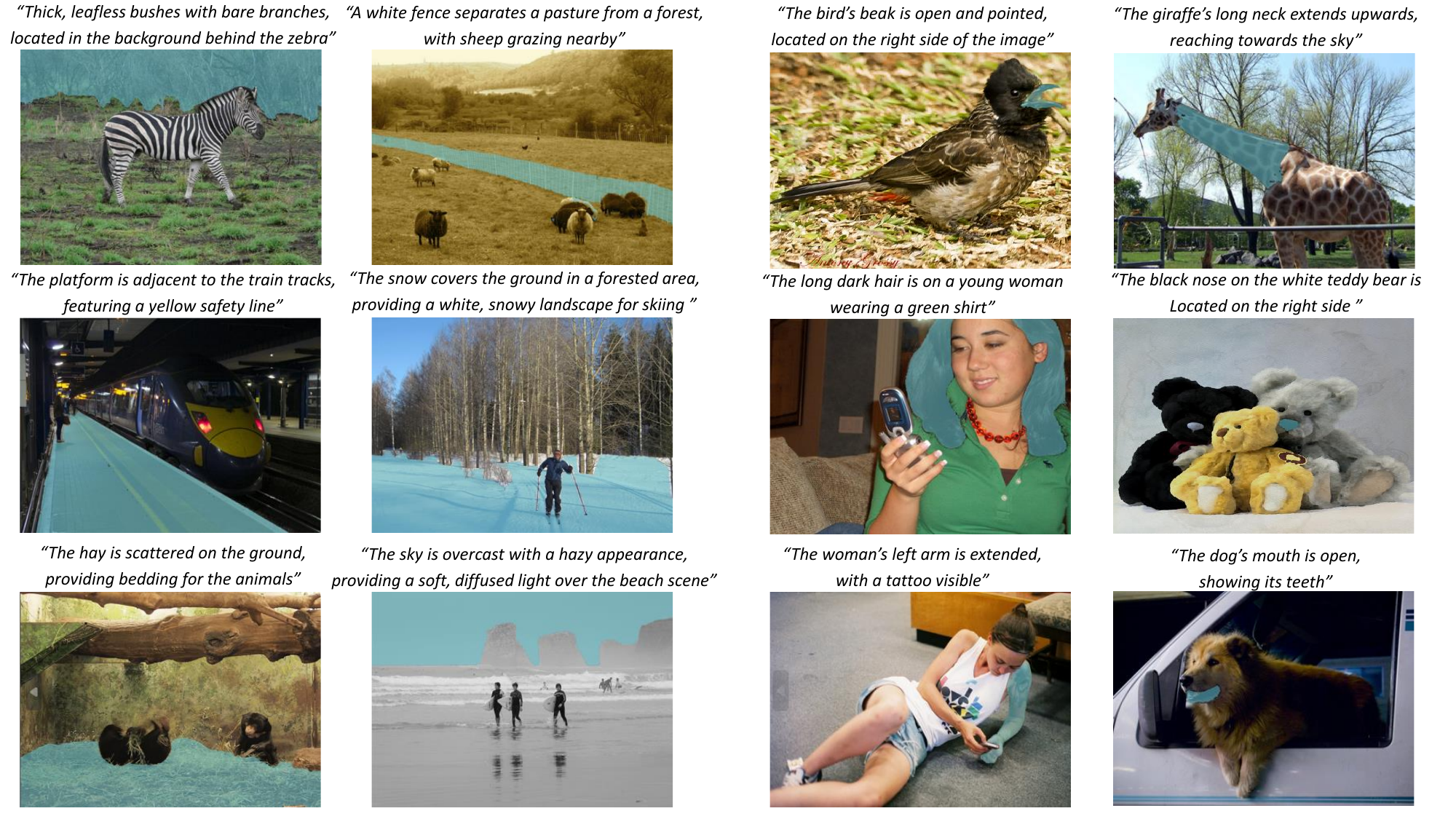}
\vspace{-.4 in}
\end{center}
   \caption{\textbf{Visualizations of samples from \gseval{} (Stuff \& Part).} We showcase some examples with \textit{stuff} and \textit{part} annotations from \gseval{}.}
\label{fig:example1}
\end{figure*}
\begin{figure*}[h]
\begin{center}
\includegraphics[width=1.0\linewidth]{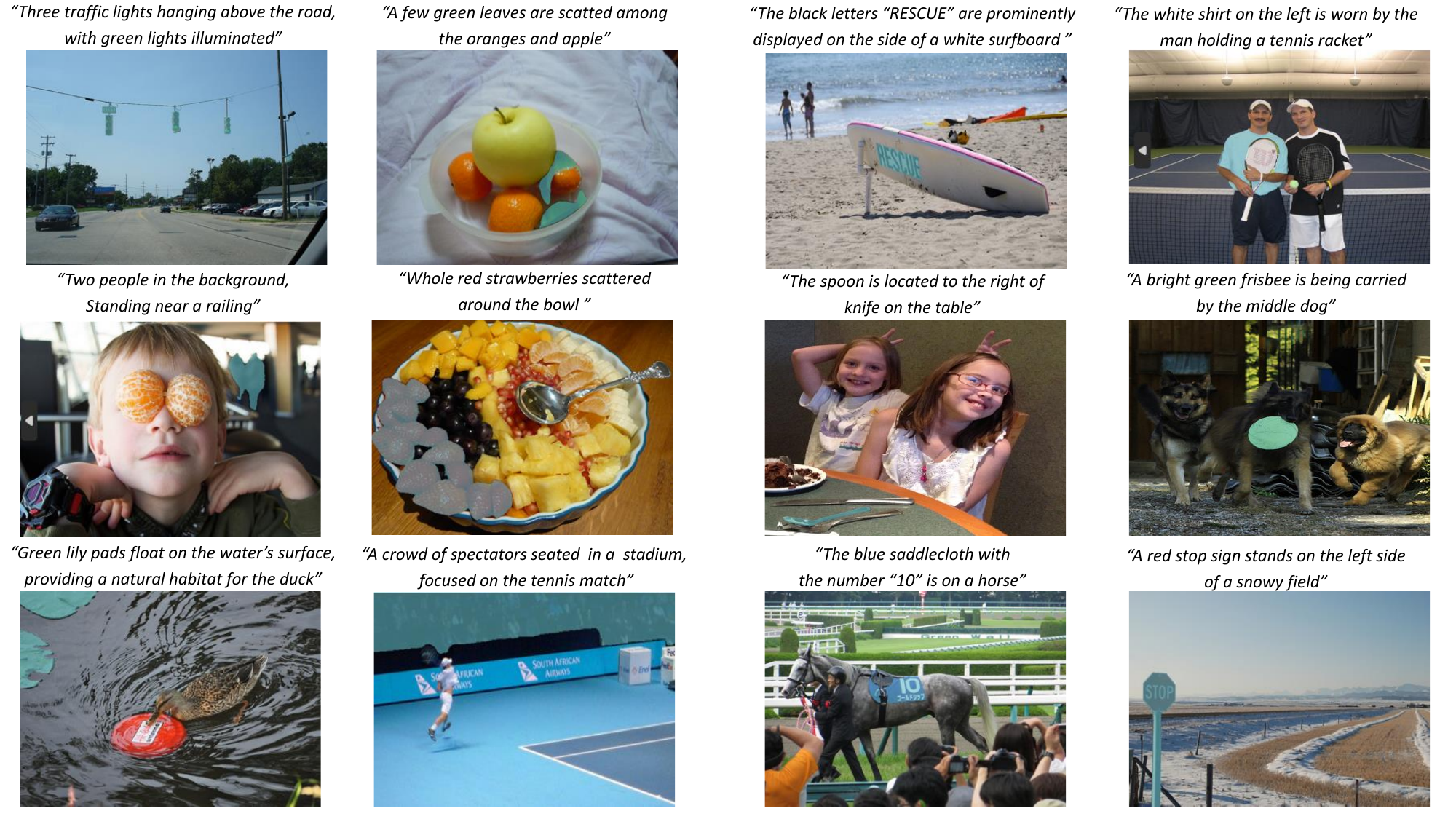}
\vspace{-.4 in}
\end{center}
   \caption{\textbf{Visualizations of samples from \gseval{} (Single \& Multi).}  We showcase some segmentation annotations of \textit{single} and \textit{multiple} objects from \gseval{}.}
\label{fig:example2}
\vspace{-.2 in}
\end{figure*}

Fig.~\ref{fig:example1} illustrates additional samples from our \gseval{}. The six images on the left represent the "stuff" class category, while the six images on the right demonstrate part-level segmentation examples.

Fig.~\ref{fig:example2} further showcases the diversity of our dataset, with the left panel displaying multi-object instances and the right panel presenting single-object examples.

\subsection{Benchmark Results on \gseval}
Tab.~\ref{tab:2} presents the zero-shot performance of previous state-of-the-art RES methods on our \gseval{} across four challenging subsets, \ie, stuff, part, multi-object and single-object, which provides a comprehensive assessment of model capabilities. 
For comparison, we also report the performance on RefCOCO/+/g benchmarks (averaged across val/testA/testB splits) using compound IoU (cIoU).

\paragraph{Limited generalization ability of specialists.}
As shown in Tab.~\ref{tab:2}, although the specialists (\eg, LAVT~\cite{lavt} and ReLA~\cite{grefcoco}) achieved good results on RefCOCO-series benchmarks, the performance on \gseval{} is poor, especially for stuff and part cases, which were not a major focus in the previous training and evaluation data. 
Compared to MLLM-based methods (multimodal large language models), it is evident that those specialists have weaker generalization capabilities. 
In addition, the dramatic performance drop highlights the limitations of existing benchmarks and validates the need for more comprehensive evaluation benchmarks like \gseval{}.

\paragraph{Limited fine-grained localization ability of MLLMs.}
As shown in Tab.~\ref{tab:2}, although some MLLM-based methods have achieved good performance on GSEval, especially for the stuff category, their performance on fine-grained part-level pixel grounding is clearly inferior to other subsets. While LLMs possess strong text understanding and reasoning capabilities, they still have significant limitations in fine-grained image localization.

Notably, we observe a significant discrepancy between the GSEval and RefCOCOs scores, such as InstructSeg~\cite{instructseg} and LISA~\cite{lisa}. InstructSeg achieves state-of-the-art performance on RefCOCOs (81.9 cIoU) but performs poorly on GSEval (52.5 gIoU). In contrast, LISA has relatively low scores on RefCOCOs (67.9 cIoU) but achieves decent accuracy on GSEval (57.6 gIoU).
We further analyze the training data of both methods and find that LISA incorporates a wide variety of datasets from different tasks including ADE20K~\cite{ade20k,ade2} and COCO-Stuff~\cite{cocostuff}, whereas InstructSeg was more focused on referring expression segmentation. This makes LISA more generalizable compared to InstructSeg and further validates that our proposed GSEval is better suited than RefCOCOs for evaluating the performance of general-purpose multi-task pixel grounding methods.

\begin{figure}
    \centering
    \includegraphics[width=\linewidth]{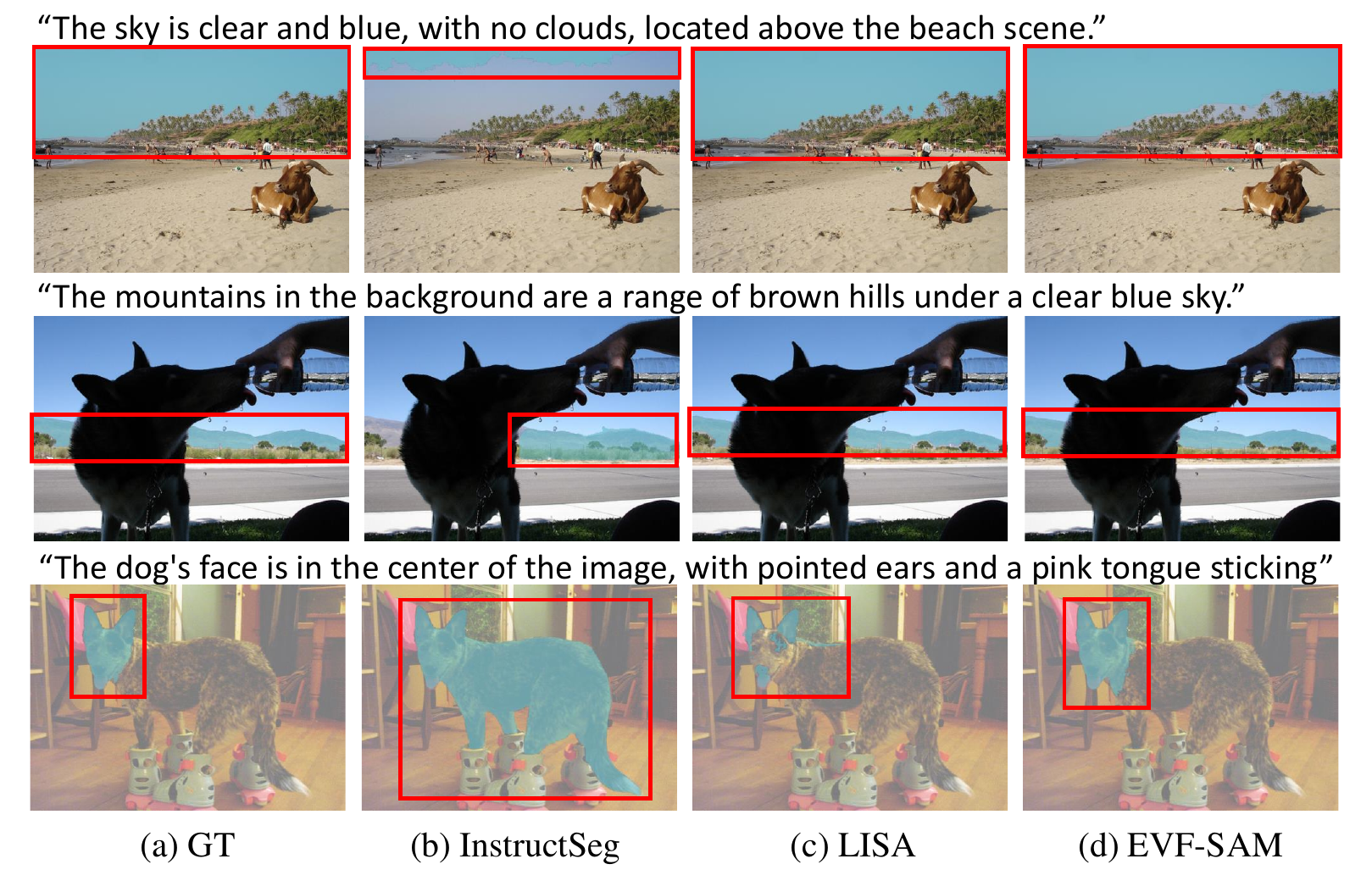}
    \caption{\textbf{The visualization comparisons of different methods on \gseval{}.} All methods are evaluated under the zero-shot setting with the public code and weights.}
    \vspace{-.1 in}
    \label{fig:gseval_zs}
\end{figure}

Moreover, we provide the visualization results on GSEval. As shown in the Fig.~\ref{fig:gseval_zs}, InstructSeg performs poorly on stuff and part categories, tending to segment entire objects instead. In contrast, LISA and EVF-SAM are better at following instructions to locate the target regions, but their segmentation accuracy still has some limitations.

\subsection{Benchmark Results on \gseval-BBox}
Additionally, we further evaluated the bbox-level grounding performance of multimodal large language models, as shown in Tab.~\ref{tab:3}. Gemini-1.5-Pro~\cite{gemini} demonstrates exceptional performance on GSEval, particularly in part-level grounding.
However, for open-source models, part-level grounding remains a challenging subset. As mentioned earlier, fine-grained object localization is relatively difficult for multimodal large language models.

\begin{figure}[t]
\begin{center}
\includegraphics[width=1.0\linewidth]{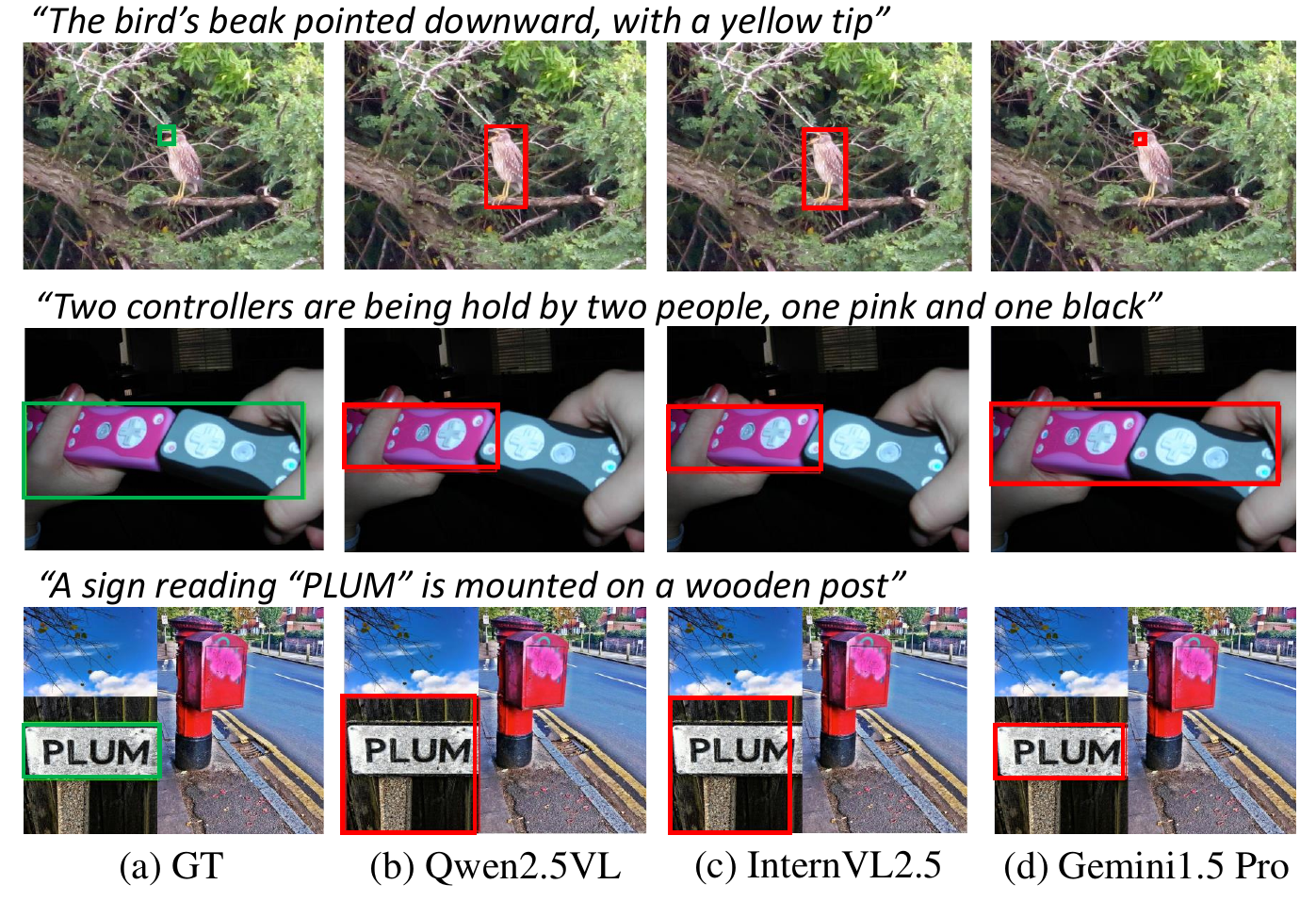}
\vspace{-30pt}
\end{center}
   \caption{\textbf{The visualization comparisons of differnet methods on our \gseval-BBox.} All open-source methods are evaluated under the zero-shot setting with the public code and weights. }
\vspace{-.12 in}
\label{fig:gseval-bbox}
\end{figure}

In Fig.~\ref{fig:gseval-bbox}, we further visualize the results of different multimodal large language models. It is clear that Qwen2.5VL~\cite{qwen2.5} and InternVL2.5~\cite{internvl2.5} struggle to locate part-level targets. Even when the targets are located, the bounding box precision is relatively low, making it difficult to achieve accurate object localization. In contrast, Gemini-1.5-Pro currently performs better, but there is still a significant gap in overall performance.

Since RefCOCO is more focused on foreground object-level grounding, with the development of multimodal large language models, it has become increasingly difficult to use RefCOCO as the primary benchmark for evaluating the performance of multimodal models. A comprehensive, multi-granularity grounding benchmark is therefore greatly needed. As a result, we believe that GSEval plays a crucial role in the grounding evaluation of multimodal models.

\begin{table*}[t]
\centering
\setlength{\tabcolsep}{15pt}
\small
\begin{tabular}{l|c|c|ccccc}
\toprule
\multirow{2}{*}{\textbf{Methods}} & \multirow{2}{*}{\textbf{Type}} & \textbf{RefCOCO/+/g} & \multicolumn{5}{c}{\textbf{\gseval}} \\
           & & (AVG) & Stuff & Part & Multi & Single & All \\
\midrule
LAVT~\cite{lavt}    &  \multirow{2}{*}{\textit{Specialist}}   & 65.8    & 6.0 & 10.7 & 45.0 & 25.8  & 22.5    \\ 
ReLA~\cite{grefcoco}  &       & 68.3    & 3.8  & 8.8  & 46.7  & 19.7  & 19.7    \\ 
\midrule
LISA-7B~\cite{lisa}     & \multirow{6}{*}{\textit{MLLM}} & 67.9 & 85.2 & 21.2 & 71.5 & 42.8  & 57.6    \\
GSVA-7B~\cite{gsva}    &    & 70.1    &76.0  &20.0 & 57.8 & 34.2  & 48.6    \\
GLaMM~\cite{GLaMM}    &    & 75.6    & \textbf{86.9} & 16.5 & 70.4 & 42.1  & 57.2    \\
PSALM~\cite{psalm}   &     & 77.1     & 39.0 & 10.0 & 53.7 & 36.9  & 37.7    \\
EVF-SAM~\cite{evfsam}  &    & 79.3 & 85.1 & 23.1 & \textbf{72.1} & \textbf{54.5}  & \textbf{62.6}    \\
InstructSeg~\cite{instructseg} & & \textbf{81.9} & 56.2 & \textbf{24.2} & 66.8 & 51.3  & 52.5 \\
\bottomrule
\end{tabular}
\caption{Comparison among previous SOTA RES methods on our \gseval{} in terms of gIoU, while we report average cIoU for RefCOCO/+/g.}
\vspace{-.06 in}
\label{tab:2}
\end{table*}

\begin{table*}
\centering
\setlength{\tabcolsep}{9pt}
\small
\begin{tabular}{l|c|c|ccccc}
\toprule
\multirow{2}{*}{\textbf{Methods}}  & \multirow{2}{*}{\textbf{Type}} & \textbf{RefCOCO/+/g} & \multicolumn{5}{c}{\textbf{\gseval-BBox}}  \\
          &   & (AVG) & Stuff & Part & Multi & Single & All\\ 
\midrule
Gemini-1.5-Pro~\cite{gemini} & \multirow{5}{*}{\parbox{2.2cm}{\centering \textit{Proprietary \\ Model}}} & -           & 86.7 & 58.5 & 61.7 & \textbf{77.3}  & 74.3      \\
Doubao-1.5-thinking-vision-pro\cite{seed1.5vl} &  & 91.3 & 86.8 & \textbf{69.7} & 83.1 & 74.3 & \textbf{79.0} \\
Doubao-1.5-vision-pro~\cite{doubao} &  & 91.6 & 80.0 & 28.8 & \textbf{85.2} & 53.3 & 64.2 \\
Claude-3.7-sonnet~\cite{claude} & &- &56.7 & 2.6 & 20.7 &9.4 &23.8\\
GPT-4o~\cite{gpt4o} & &- &42.2 &2.0 &15.3 &6.1 &17.3 \\
\midrule
InternVL3-78B\cite{zhu2025internvl3}& \multirow{10}{*}{\parbox{2.2cm}{\centering \textit{Open-source \\ Model}}} &  91.4          & 69.3 & 13.8 & 61.1 & 49.5  & 52.9      \\
InternVL3-8B\cite{zhu2025internvl3}& &  89.6          & 77.3 & 8.4 & 56.6 & 46.5  & 52.3      \\
InternVL2.5-78B~\cite{internvl2.5}&  &  92.3          & 85.3 & 16.8 & 63.2 & 55.7  & 62.2      \\
InternVL2.5-8B~\cite{internvl2.5}& &  87.6          & 91.3 & 7.3 & 65.7 & 47.3  & 58.2      \\
Qwen2.5-VL-72B~\cite{qwen2.5}& & 90.3         & 88.4 & 31.2 & 42.8 & 64.6  & 62.5       \\
Qwen2.5-VL-7B~\cite{qwen2.5}& & 86.6             & \textbf{93.0} & 17.6 & 75.9 & 59.1  & 66.7     \\
DeepSeek-VL2~\cite{deepseekvl2}& & \textbf{93.0}      & 86.5 & 12.7 & 64.8 & 51.2  & 60.8    \\
Ferret-13B~\cite{ferret}& &85.6       & 80.6 & 21.3 & 58.0 & 46.6  & 55.1 \\
Ferret-7B~\cite{ferret}& &83.9       & 82.1 & 17.6 & 56.0 & 43.2  & 53.3 \\
Mistral-Small-3.1-24B~\cite{mistral}& & -      & 16.0 & 3.5 & 20.3 & 10.7  & 13.2 \\
\bottomrule
\end{tabular}
\caption{Comparison among previous grounding methods on our \gseval-BBox. All metrics measure referring expression comprehension accuracy (\%).}
\vspace{-.12 in}
\label{tab:3}
\end{table*}
\section{Experiments}
In this section, we conduct a series of ablation studies to further validate the effectiveness of the automatically annotated training dataset, ~\gstrain{}. 
We describe the details of each experiment in the corresponding subsection.

\subsection{Effectiveness on Pixel Grounding Methods}
To evaluate the impact of our \gstrain{}, we select two representative methods, \ie, EVF-SAM~\cite{evfsam} and LISA-7B~\cite{lisa}, and test them with and without our \gstrain{}.
\paragraph{Experimental details.}
For the baseline, we followed the original settings of EVF-SAM and LISA-7B. Specifically, EVF-SAM utilizes multiple datasets, including the RefCOCO series \cite{referit,refcoco,refcocog,refcocog2}, Objects365 \cite{objects365}, ADE20K \cite{ade20k,ade2}, Pascal-Part \cite{pascal_part}, HumanParsing \cite{liang2015human}, and PartImageNet \cite{he2022partimagenet}.

Meanwhile, LISA-7B employs the RefCOCO series \cite{referit,refcoco,refcocog,refcocog2}, ADE20K \cite{ade20k,ade2}, COCO-Stuff \cite{cocostuff}, PACO-LVIS \cite{ramanathan2023paco}, PartImageNet \cite{he2022partimagenet}, Pascal-Part \cite{pascal_part}, as well as LLaVAInstruct-150k for LLaVA-v1 \cite{llavav1}, LLaVA-v1.5-mix665k for LLaVA v1.5 \cite{llavav1.5}, and ReasonSeg \cite{lisa}.

\begin{table}
\centering
\small
\setlength{\tabcolsep}{2pt}
\begin{tabular}{l|c|lll}
\toprule
\textbf{Methods} & \textbf{GSTrain-10M} & \textbf{\gseval{}} & \textbf{gRefCOCO} & \textbf{RefCOCOm} \\ 
\midrule
EVF-SAM&            & 62.6 & 63.5 & 51.3 \\
EVF-SAM& \checkmark & 77.3 \resizebox{!}{0.7\height}{\textcolor{red}{+14.7}} & 66.4 \resizebox{!}{0.7\height}{\textcolor{red}{+2.9}}& 55.3 \resizebox{!}{0.7\height}{\textcolor{red}{+4.0}}\\
\hline
LISA-7B&            & 57.6 & 32.2 & 34.2 \\
LISA-7B& \checkmark & 73.6 \resizebox{!}{0.7\height}{\textcolor{red}{+16.0}}& 36.3 \resizebox{!}{0.7\height}{\textcolor{red}{+4.1}}& 39.3 \resizebox{!}{0.7\height}{\textcolor{red}{+5.1}} \\
\bottomrule
\end{tabular}
\caption{Ablation study on the effectiveness of our dataset on different methods.}
\vspace{-.3 in}
\label{table:4}
\end{table}

\paragraph{Results.}
In Tab.~\ref{table:4}, we evaluate the performance using the proposed \gstrain{} of EVF-SAM and LISA on several benchmarks.
The results demonstrate that our dataset brings significant performance improvements to both methods across all benchmarks. 

For EVF-SAM, 
we observe 14.7, 2.9, and 4.0 points improvements on \gseval{}, gRefCOCO, and RefCOCOm benchmarks, respectively.
The consistent gains across different evaluation benchmarks highlight the transferability of knowledge acquired from our diverse training corpus.

For LISA-7B,
we observe 16.0, 4.1, and 5.1 points improvements on \gseval{}, gRefCOCO, and RefCOCOm benchmarks, respectively.
The greater relative improvement observed in LISA-7B suggests that language-centric models particularly benefit from our dataset's text diversity and the richness of our referring expressions (averaging 16 words in length).

Furthermore, the consistent improvements across both external benchmarks (gRefCOCO and RefCOCOm) demonstrate that the knowledge gained from our dataset generalizes well beyond the specific distribution of \gseval. 
Considering the gRefCOCO focuses on multi-object relationships, while RefCOCOm evaluates fine-grained semantic understanding,
the performance gains across these diverse evaluation scenarios demonstrate the robustness and generality of the proposed GSTrain-10M.

These results collectively validate that the proposed GSSculpt produces high-quality training data that enhances model performance across architectural paradigms, supporting more accurate and nuanced referring expression segmentation capabilities.

\subsection{Comparisons with Other Automated Datasets}

To further validate our approach, we conduct a comparative analysis between the proposed \gstrain{} and GranD, 
another state-of-the-art automatically generated dataset introduced with GLaMM~\cite{GLaMM}. 
\paragraph{Experimental details.}
For fair comparisons, we randomly selected 100k samples from both GranD and the proposed \gstrain{} to train models under the LAVT, EVF-SAM and LISA framework.

\paragraph{Results.}

 As evidenced in Tab.~\ref{table:5}, our proposed GSTrain dataset consistently outperforms the GranD dataset across all evaluated models and benchmarks. This benefit is clear for the specialist model, LAVT, which sees performance boosts of 2.2, 1.1, and 2.0 percentage points on RefCOCO, gRefCOCO, and RefCOCOm, respectively. The advantages of GSTrain are even more pronounced for the MLLM-based models. EVF-SAM, for instance, achieves gains of 2.8, 0.9, and 1.2 points across the same benchmarks. The most significant improvement is observed with LISA-7B, which not only improves by 2.5 points on RefCOCO but also exhibits a substantial 8.0-point increase on RefCOCOm (37.6 vs. 29.6). This considerable performance gain on a benchmark known for its linguistic complexity underscores that GSTrain provides the rich, high-quality data crucial for enabling advanced models to better interpret fine-grained textual descriptions.

\begin{table}[t]
\centering
\small
\setlength{\tabcolsep}{3pt}
\begin{tabular}{l|c|lll}
\toprule
\textbf{Methods} & \textbf{Dataset} & \textbf{RefCOCO} & \textbf{gRefCOCO} & \textbf{RefCOCOm} \\ 
\midrule
LAVT&   GranD         & 39.5 &26.3  &27.1  \\
LAVT& GSTrain & 41.7 \resizebox{!}{0.7\height}{\textcolor{red}{+1.2}} &27.4 \resizebox{!}{0.7\height}{\textcolor{red}{+1.1}} &29.1 \resizebox{!}{0.7\height}{\textcolor{red}{+2.0}}   \\
\midrule
EVF-SAM&   GranD         & 60.7 & 28.0 & 31.5 \\
EVF-SAM& GSTrain & 63.5 \resizebox{!}{0.7\height}{\textcolor{red}{+2.8}} & 28.9 \resizebox{!}{0.7\height}{\textcolor{red}{+0.9}}& 32.7 \resizebox{!}{0.7\height}{\textcolor{red}{+1.2}}\\
\midrule
LISA-7B&  GranD          & 63.1 & 29.5 & 29.6 \\
LISA-7B& GSTrain & 65.6 \resizebox{!}{0.7\height}{\textcolor{red}{+2.5}}& 31.3 \resizebox{!}{0.7\height}{\textcolor{red}{+1.8}}& 37.6 \resizebox{!}{0.7\height}{\textcolor{red}{+8.0}} \\
\bottomrule
\end{tabular}
\caption{Ablation study on the effectiveness of different automated datasets.}
\vspace{-.1 in}
\label{table:5}
\end{table}

\begin{table}[t]
\centering
\small
\setlength{\tabcolsep}{12pt} 
\label{tab:refcocog_expression_length_results} 
{\begin{tabular}{@{}llll@{}} 
\toprule
Expression Length & $\leq$10 & 11 - 15 & $\geq$16 \\
\midrule
w/o GSTrain-10M & 79.2 & 77.4 & 81.7 \\
w/~~ GSTrain-10M   & 79.7 \resizebox{!}{0.7\height}{\textcolor{red}{+0.5}} & 78.0 \resizebox{!}{0.7\height}{\textcolor{red}{+0.6}} & 82.8 \resizebox{!}{0.7\height}{\textcolor{red}{+1.1}} \\
\bottomrule
\end{tabular}}
\caption{Performance comparison on RefCOCOg across different referring expression lengths.}
\vspace{-.1 in}
\label{tab:refcocog_expression_length_results}
\end{table}

\begin{table}[t]
\centering
\small
\setlength{\tabcolsep}{1pt}
{\begin{tabular}{l|lll}
\toprule
\text{Filter Type} & \textbf{RefCOCO} & \textbf{gRefCOCO} & \textbf{RefCOCOm} \\ 
\midrule
No Filter        & 36.9 & 26.2 & 25.9 \\
CLIP-based Filter         & 60.7 & 28.0 & 31.5 \\
IoU-based Filter (\textit{\text{Ours.}}) & 63.5 \resizebox{!}{0.7\height}{\textcolor{red}{+2.8}} & 28.9 \resizebox{!}{0.7\height}{\textcolor{red}{+0.9}}& 32.7 \resizebox{!}{0.7\height}{\textcolor{red}{+1.2}}\\
\bottomrule
\end{tabular}}
\caption{Ablation study on the effectiveness of different filter types.}
\vspace{-.2 in}
\label{tab:filter}
\end{table}

\subsection{Analysis of long expressions}
Our quantitative results (Tab.~\ref{tab:refcocog_expression_length_results}) demonstrate that training the EVF-SAM model with our GSTrain-10M dataset yields a significant 1.1 cIoU point improvement on RefCOCOg for expressions of 16 words or more. We attribute this to the textual diversity and complexity of GSTrain-10M. More samples are visualized in revision.

\subsection{Analysis of different filter types}
We compared different filtering methods: no filter (baseline for unfiltered data), a CLIP-based filter (GranD), and our proposed IoU-based filter. The results in Tab.~\ref{tab:filter} indicate that filtering is essential for generating high-quality data. Significantly, our IoU-based filter demonstrates superior performance over the CLIP-based approach, achieving a more substantial reduction in labeling noise and enhancing data precision.

\subsection{Scalability}
To investigate the effect of data scalability on model performance, 
we train EVF-SAM with different proportions of our \gstrain{} without other datasets, such as RefCOCO.
\paragraph{Experimental details.}
We follow the training settings from \cite{evfsam} and train EVF-SAM with different proportions of our \gstrain{} (0\%, 20\%, 50\%, 100\%) for 10 epochs and then evaluate it on the proposed \gseval{}.

\paragraph{Results.}
As shown in Tab.~\ref{table:6}, the results show that model performance increases significantly across all aspects as the data proportion increases.
Notably, we can observe a clear rising curve with the increment of training data ratio, demonstrating the scalability of the proposed \gstrain{}.

\begin{table}
\centering
\small
\setlength{\tabcolsep}{9pt}
\begin{tabular}{l|cccc|c}
\toprule
\multirow{2}{*}{\textbf{Ratio}} & \multicolumn{5}{c}{\textbf{\gseval}}\\
           & Stuff & Part & Multi & Single & All\\ 
\midrule
0\%         & 85.1 & 23.1 & 72.1 & 54.5  & 62.6\\
20\%        & 93.2 & 43.2 & 85.0 & 68.2  & 75.4     \\
50\%           & 94.6 & 54.0 & 88.3 & 72.7  & 79.6    \\
100\%       & 94.7 & 59.2 & 89.0 & 72.4  & 80.3     \\
\bottomrule
\end{tabular}
\caption{Ablations on the scalability.}
\vspace{-.1 in}
\label{table:6}
\end{table}

\section{Conclusion}
In this paper, we present GroundingSuite, which
contains a vision-language model (VLM) based automatic annotation framework,
a large-scale, textually diverse training corpus (9.56M masks), 
and a comprehensive evaluation framework. 
Extensive experiments demonstrate that models trained on GSTrain-10M consistently establish new state-of-the-art results across multiple benchmarks. 
The proposed GroundingSuite lays a solid foundation for the visual language understanding domain, providing valuable support for future research endeavors.

{
    \small
    \bibliographystyle{ieeenat_fullname}
    \bibliography{main}
}
\clearpage
\maketitlesupplementary


\appendix

\section{Dataset Details}
\begin{figure}[h]
\begin{center}
\includegraphics[width=1.0\linewidth]{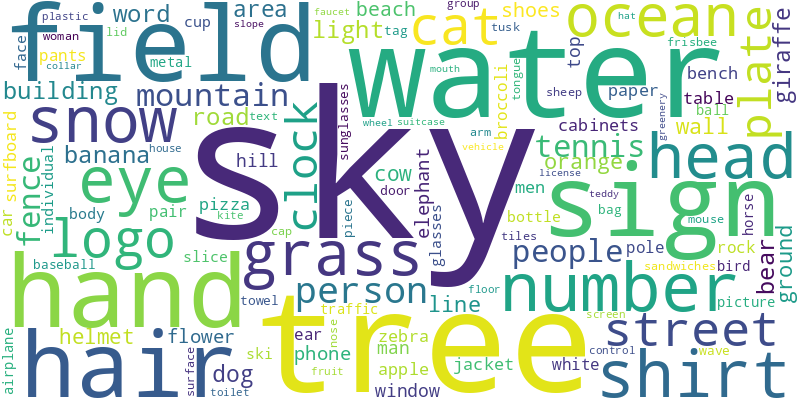}
\end{center}
   \caption{The word cloud of \gseval.}
\label{fig:wordcloud}
\end{figure}

Fig.~\ref{fig:wordcloud} shows the word cloud visualization of our benchmark's textual descriptions, highlighting the linguistic diversity and domain coverage of \gseval{}.

\vspace{-5 pt}

\begin{figure}[h]
\begin{center}
\includegraphics[width=1.0\linewidth]{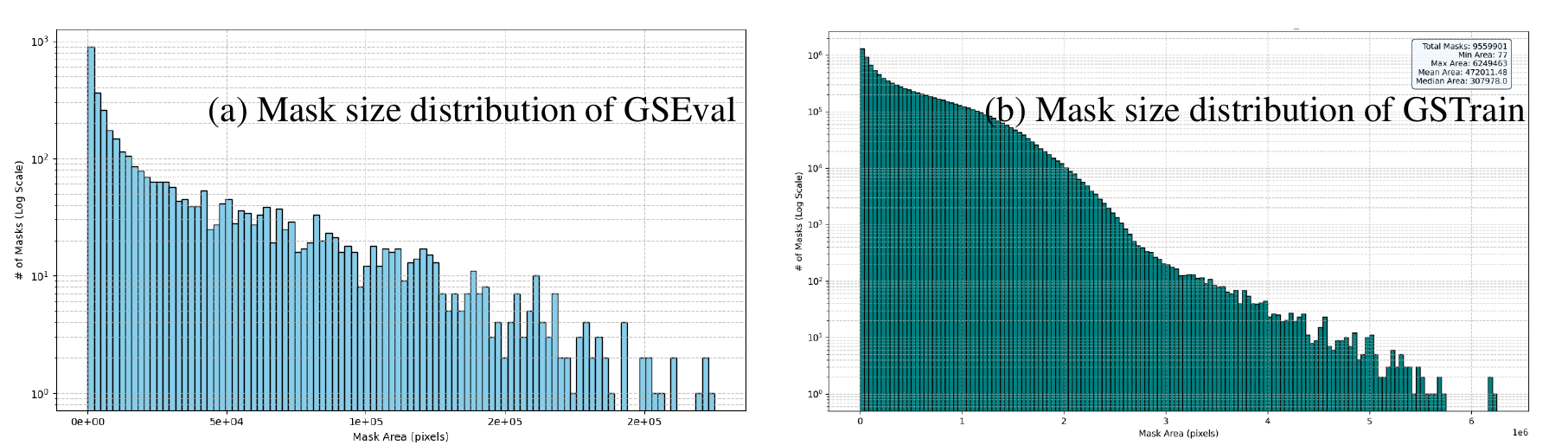}
\end{center}
\vspace{-.3 in}
\caption{Distribution of mask sizes of GroundingSuite (zoom in for more detailed information).}
\label{fig:mask_distribution}
\end{figure}

\begin{figure}[h]
\begin{center}
\includegraphics[width=1.0\linewidth]{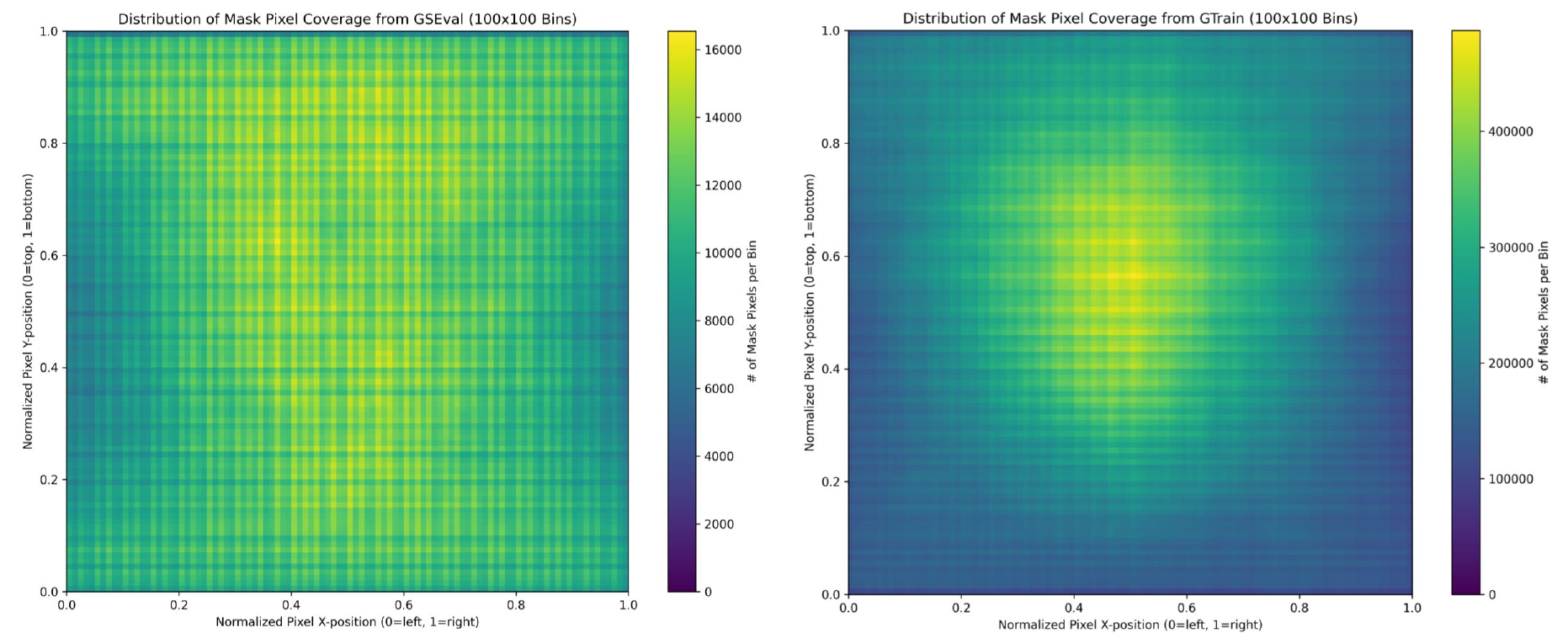}
\end{center}
\vspace{-.3 in}
\caption{Distribution of mask positions of GroundingSuite (zoom in for more detailed information).}
\label{fig:position_distribution}
\end{figure}

\begin{figure}[h]
\begin{center}
\includegraphics[width=1.0\linewidth]{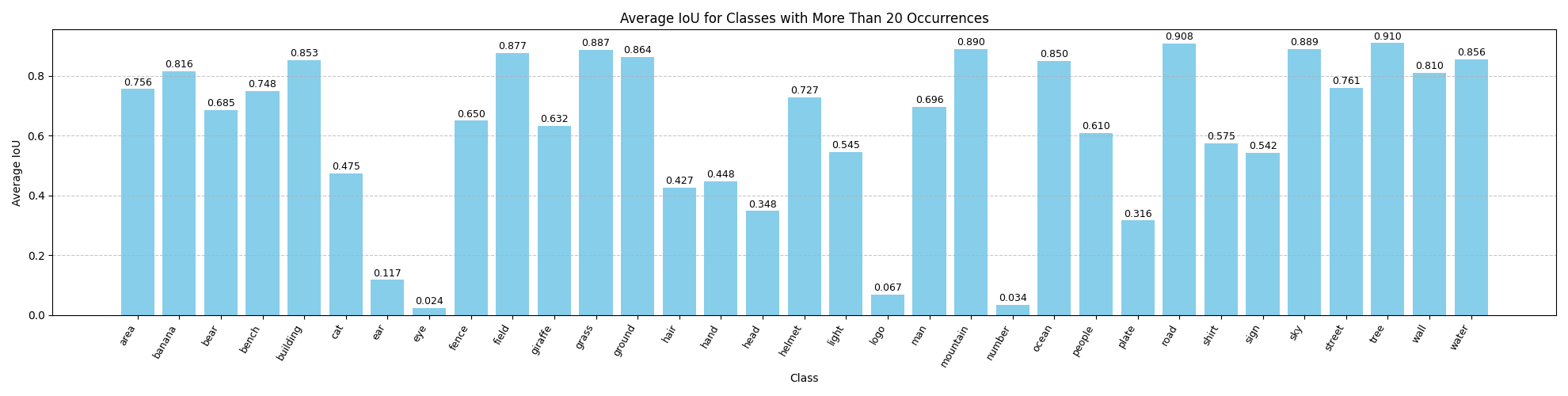}
\end{center}
\vspace{-.3 in}
\caption{Partial statistics of fine-grained class IoU (zoom in for more detailed information).}
\label{fig:class}
\end{figure}

In addition, we analyze mask size distribution (Fig.~\ref{fig:mask_distribution}) and position distribution (Fig.~\ref{fig:position_distribution}) towards understanding mask size and position biases, which demonstrate GSTrain (Sampled from SA-1B) and GSEval (Sampled from unlabeled COCO) have significantly different distributions. We add a detailed breakdown of performance for different object classes. Partial statistics are presented in Fig.~\ref{fig:class}.

\section{Prompt}
\subsection{Global caption generation}
We utilize the InternVL2.5-78B model to generate comprehensive global captions for the images. The specifically designed prompt template employed for this purpose is presented in Figure~\ref{fig:prompt1}.

\subsection{Grounding text generation}
\begin{figure*}[t]
  \begin{center}
    \includegraphics[width=1.0\linewidth]{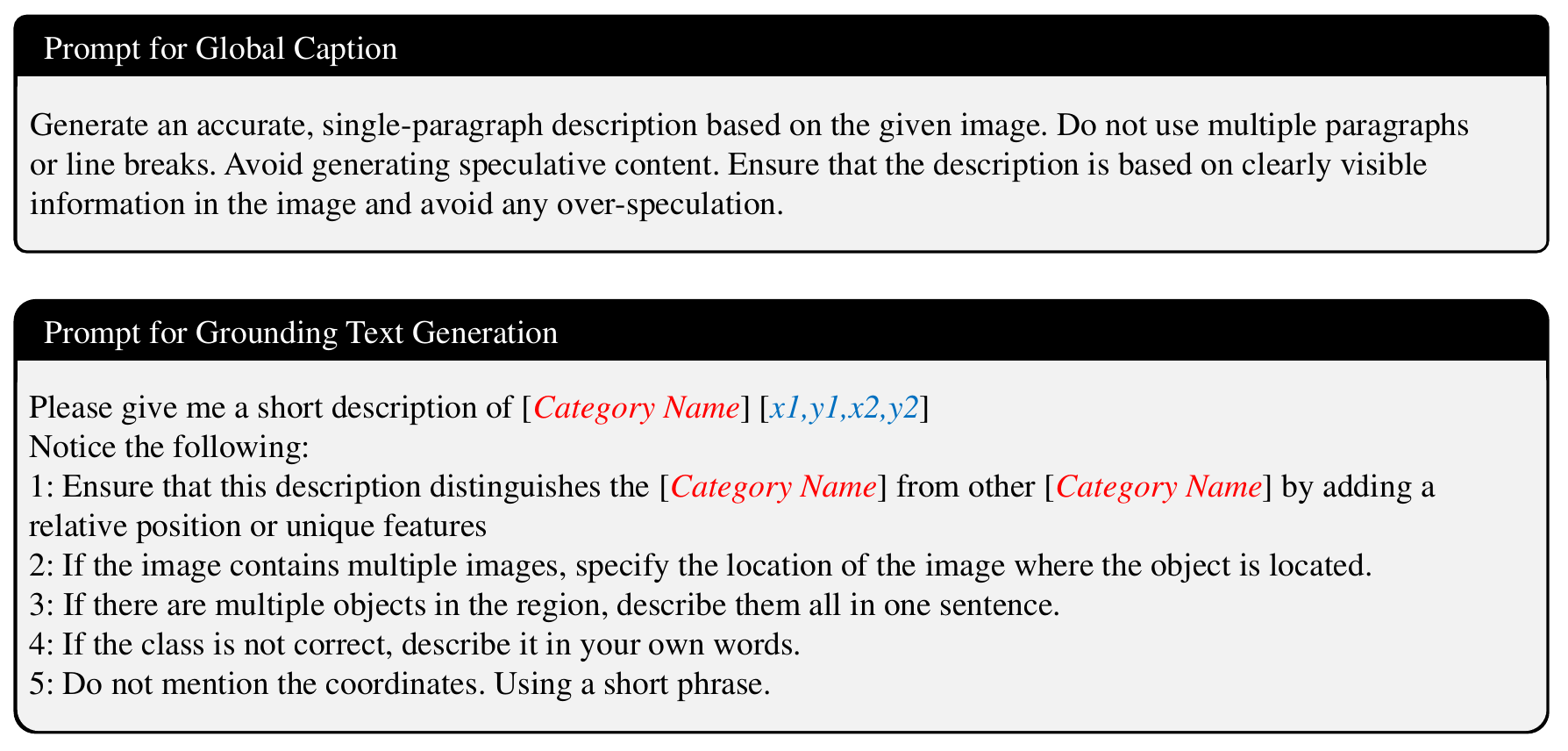}
  \end{center}
  \caption{Prompt for global caption and grounding text generation}
  \label{fig:prompt1}
\end{figure*}

We employ specialized prompt templates with InternVL2.5 to generate unambiguous references that emphasize spatial relationships and distinctive visual features. The detailed prompt template is illustrated in Figure~\ref{fig:prompt1}.

\subsection{Noise filtering}
\begin{figure*}[t]
\begin{center}
\includegraphics[width=1.0\linewidth]{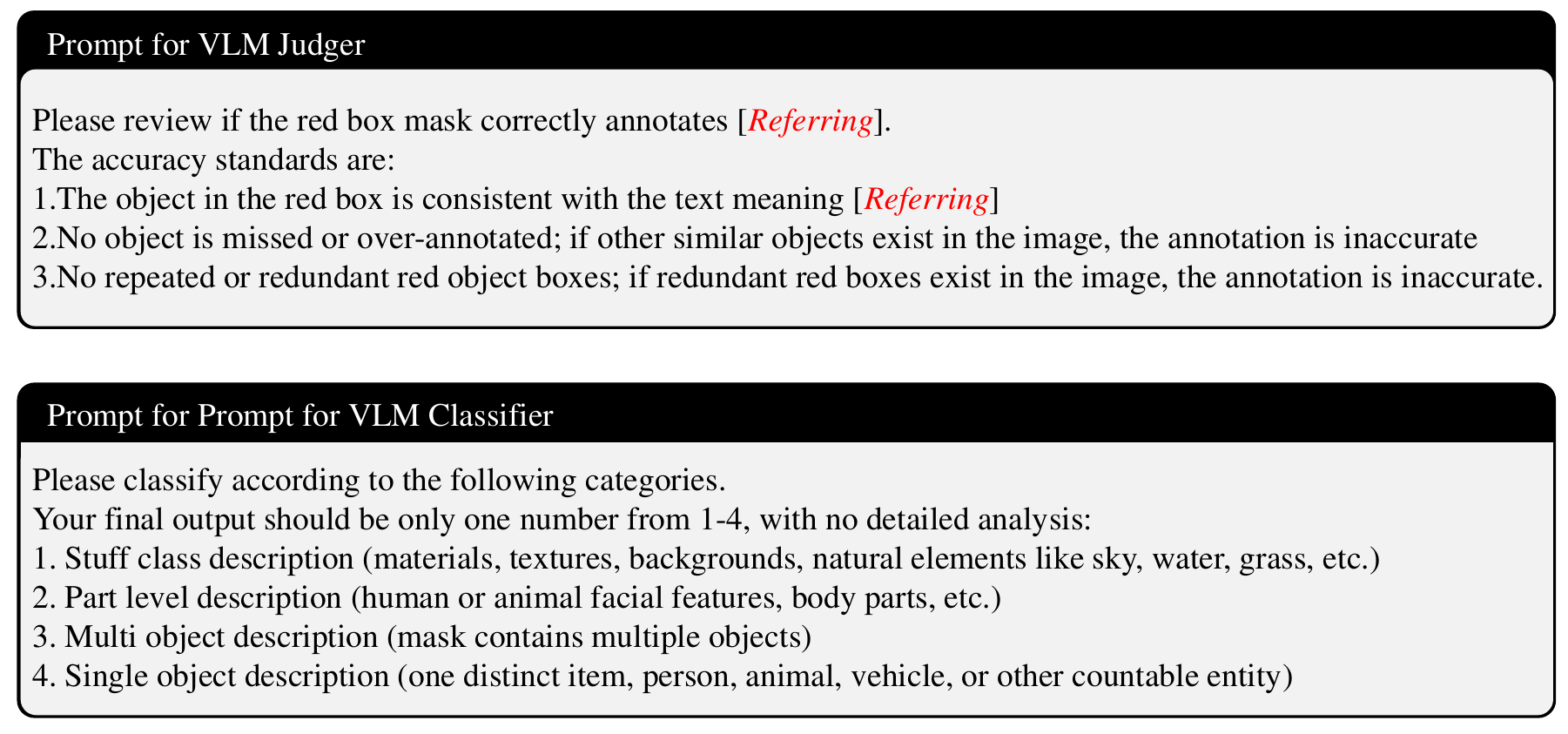}
\end{center}
\caption{Prompt for noise filtering}
\label{fig:prompt2}
\end{figure*}

During the noise filtering stage, we employ a two-step approach: first prompting the Vision-Language Model (VLM) to assess referring expression accuracy, then using specialized prompts to classify the referring expressions by category. Both prompt templates are illustrated in Figure~\ref{fig:prompt2}.

\subsection{Prompt for different grounding models}

For different grounding models, we apply customized prompt templates to generate bounding box coordinates:

\begin{itemize}
    \item \textbf{Gemini-1.5-Pro:} \\
    Return a bounding box for [\textit{\textbf{\textcolor{red}{Referring}}}] in this image in [xmin, ymin, xmax, ymax] format.
    
    \item \textbf{GPT-4o and Claude-3.7-sonnet:} \\
    In this image, please locate the object described as: '[\textit{\textbf{\textcolor{red}{Referring}}}]'. Provide the bounding box coordinates in the format [x\_min, y\_min, x\_max, y\_max]. You can use either absolute pixel coordinates or normalized coordinates (0-1 range).
    
    \item \textbf{Doubao-1.5-vision-pro:} \\
    Please provide the bounding box coordinate of the region this sentence describes: [\textit{\textbf{\textcolor{red}{Referring}}}]
    
    \item \textbf{InternVL2.5:} \\
    Please provide the bounding box coordinate of the region this sentence describes: \texttt{"<ref>}[\textit{\textbf{\textcolor{red}{Referring}}}]\texttt{</ref>"}
    
    \item \textbf{Qwen2.5-VL:} \\
    Please provide the bounding box coordinate of the region this sentence describes: \texttt{<|object\_ref\_start|>}[\textit{\textbf{\textcolor{red}{Referring}}}]\texttt{<|object\_ref\_end|>}
    
    \item \textbf{Deepseek-VL-2:} \\
    \texttt{<image><|ref|>}[\textit{\textbf{\textcolor{red}{Referring}}}]\texttt{<|/ref|>}.
\end{itemize}

\end{document}